\pdfoutput=1

\documentclass[11pt]{article}

\usepackage[final]{emnlp2021}

\usepackage{times}
\usepackage{latexsym}

\usepackage[T1]{fontenc}

\usepackage[utf8]{inputenc}

\usepackage{microtype}
\usepackage{graphicx}
\usepackage{amsmath}
\usepackage{hyperref}
\usepackage[colorinlistoftodos]{todonotes}
\usepackage{multirow}
\usepackage{mathtools}
\title{SupCL-Seq: Supervised Contrastive Learning for\\Downstream Optimized Sequence Representations}

\author{
  Hooman Sedghamiz$^1$ \quad
  Shivam Raval$^1$ \quad
  \textbf{Enrico Santus}$^1$ \quad
  \textbf{Tuka Alhanai}$^2$ \quad \\
  \textbf{Mohammad Ghassemi}$^3$ \\
  $^1$ DSIG - Bayer Pharmaceuticals, New Jersey, USA \\
  $^2$ New York University, Abu Dhabi, UAE \\
  $^3$ Michigan State University, Michigan, USA \\
  \texttt{hooman.sedghamiz@bayer.com}, \texttt{sbraval@asu.edu},\\
  \texttt{enrico.santus@bayer.com}, \texttt{tuka.alhanai@nyu.edu}, \texttt{ghassem3@msu.edu}
}

\begin{document}
\maketitle
\begin{abstract}
While contrastive learning is proven to be an effective training strategy in computer vision, Natural Language Processing (NLP) is only recently adopting it as a \textit{self-supervised} alternative to Masked Language Modeling (MLM) for improving sequence representations. 
This paper introduces \textit{SupCL-Seq}, which extends the \textit{supervised} contrastive learning from computer vision to the optimization of sequence representations in NLP. 
By altering the \textit{dropout} mask probability in standard Transformer architectures (e.g.~BERT\textsubscript{base}), for every representation (\textit{anchor}), we generate augmented altered \textit{views}. A supervised contrastive loss is then utilized to maximize the system's capability of pulling together similar samples (e.g., \textit{anchors} and their altered \textit{views}) and pushing apart the samples belonging to the other classes.
Despite its simplicity, \textit{SupCL-Seq} leads to large gains in many sequence classification tasks on the GLUE benchmark compared to a standard BERT\textsubscript{base}, including $6 \%$ absolute improvement on CoLA, $5.4 \%$ on MRPC, $4.7 \%$ on RTE and $2.6 \%$ on STS-B. We also show consistent gains over \textit{self-supervised} contrastively learned representations, especially in non-semantic tasks. Finally we show that these gains are not solely due to augmentation, but rather to a downstream optimized sequence representation. Code: \href{https://github.com/hooman650/SupCL-Seq}{https://github.com/hooman650/SupCL-Seq}
\end{abstract}

\section{Introduction}
Sequence classification is a fundamental problem in natural language processing (NLP), as it has a wide range of applications, including but not limited to the tasks such as sentiment analysis, inference and question answering~\cite{DBLP:journals/corr/abs-2004-03705}. Cross-entropy loss is generally the default loss function in training neural networks for NLP downstream tasks~\citep{DBLP:journals/corr/abs-1805-07836, c51d68a3106242f08ed001d0c46320b3}. Recently, thanks to the simplicity of augmentation methods in computer vision (e.g., zooming, cropping, rotation, etc.), \textit{self-supervised} and \textit{supervised} variants of contrastive learning proved to be effective training approaches in image classification tasks~\citep{Wu_2018_CVPR, Oliver2019, khosla2020}. These methods aim at optimizing the representations by minimizing the distance between similar samples and maximizing it between diverse samples~\cite{DBLP:journals/corr/abs-2006-10029}.
\citet{Gao2021} proposed to leverage the built-in \textit{dropout} masks in attention and fully-connected layers of Transformers~\cite{vaswani2017} to introduce noise in the embedding representations. This is obtained by simply passing \textit{twice} the same input and using different dropout masks. In this way, for every representation (\textit{anchor}), altered \textit{views} are generated. \citet{Gao2021} applied this augmentation approach to improve the semantic representation of a sequence in a \textit{self-supervised} fashion, by taking an input sentence and contrasting its similarity against its augmented version and the remaining samples in a batch. The authors further extended this approach by employing positive (i.e., entailment) and negative (i.e., contradiction) examples from natural language inference (NLI) datasets. The resulting sentence embeddings achieved large gains in semantic textual similarity (STS) tasks.

\begin{figure*}[!ht]
\centering
\includegraphics[width=0.9\textwidth]{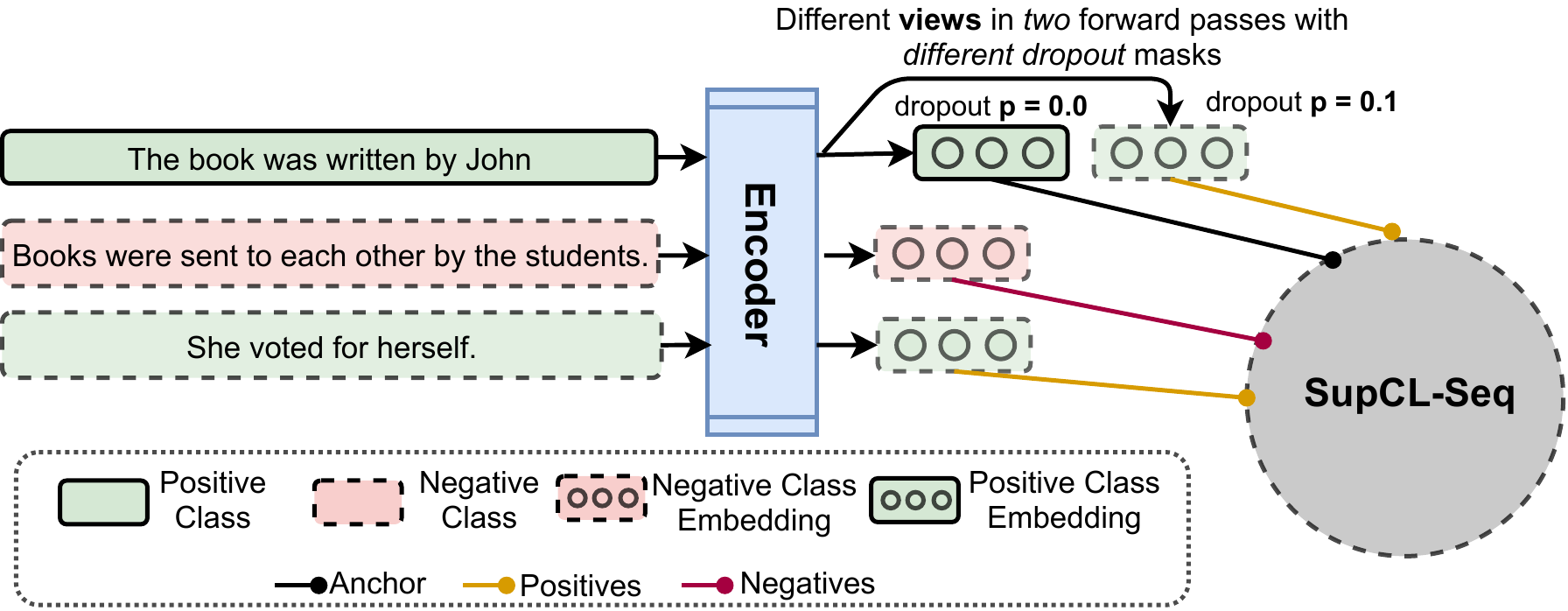}
\caption{\textit{SupCL-Seq} applied to COLA~\cite{warstadt2018neural}. \textit{SupCL-Seq} first forward propagates the input N times (in this example N = 2) through the same encoder (e.g.~BERT\textsubscript{base}) with N different dropout masks (e.g. $p=0$ and $p=0.1$ respectively) and obtains their corresponding noisy embedding views. The noisy embeddings that belong to the same class are then employed as the positive pairs for the original input (\textit{anchor} with dropout mask of $p=0$). In this way, the samples belonging to the \textit{negative} class effectively are used as negatives.}
\label{fig:SupCSE} 
\end{figure*}

To the best of our knowledge, however, contrastive learning has not yet been applied in a supervised fashion to optimize sequence representations towards downstream tasks. \footnote{During the review process, we were made aware of a contemporaneous work by \citet{gunel2020supervised} on supervised contrastive learning for natural language processing. A major difference between their work and ours lays in the adopted augmentation methodology.} Inspired by the recently proposed \textit{supervised} contrastive learning in computer vision~\cite{khosla2020}, in this paper we introduce \textit{SupCL-Seq}, which extends the \textit{self-supervised} contrastive method by \citet{Gao2021} to a \textit{supervised} contrastive learning approach, in which \textit{anchors} and altered \textit{views}, along with their classification labels, are used to learn downstream optimized sequence representations by means of contrastive learning (~see Figure~\ref{fig:SupCSE} for details).


\textit{SupCL-Seq} is simple and can be applied to any sequence classification task, on any arbitrary number of classes. In our experiments, \textit{SupCL-Seq} leads to large gains in several tasks of the GLUE benchmark \citep{wang-etal-2018-glue}, including CoLA ($6 \%$ Matthew`s correlation coefficient absolute improvement), MRPC ($5.4 \%$ accuracy score absolute improvement), RTE ($4.3 \%$ accuracy score absolute improvement) and STS-B ($2.6 \%$  Spearman's rank correlation coefficient absolute improvement).

The main contributions of this paper are:

\begin{itemize}
    \item The adaptation from computer vision to NLP of a \textit{supervised} contrastive learning approach for sequence classification tasks (\textit{SupCL-Seq}), which extends \citet{Gao2021}'s approach by optimizing the sequence representations for any downstream task, independently on the number of classes.
    \item Empirical demonstration that \textit{SupCL-Seq} leads to significant gains in many text classification tasks in GLUE~\cite{wang-etal-2018-glue} using a standard transformer such as BERT\textsubscript{base} \cite{BERT2018}.
\end{itemize}


\section{Method}
\textit{SupCL-Seq} extends the self-supervised contrastive learning~\cite{Gao2021} for improving semantic representations to a supervised setting, in which representations are optimized towards the downstream task, independently on the number of labels.

The augmentation step is obtained by forward propagating the input batch $N$ times in the same encoder with $N$ distinct \textit{dropout} masks (i.e., different dropout probabilities). The generated altered \textit{views}, along with their \textit{anchor}'s label, are then used to optimize the sequence representations through a supervised contrastive loss function~\cite{khosla2020}. Figure~\ref{fig:SupCSE} details our training approach. 

Formally, our pipeline consists of a single \textit{Encoder Transformer}, \textit{$Enc(.)$} (i.e., BERT\textsubscript{base} with $\approx$110M parameters \citep{BERT2018}). This encoder generates $N$ altered embeddings, $\mathbf{\tilde{x}_n} = Enc(\mathbf{x}, p_n)$, for each input $\mathbf{x}$ and \textit{dropout} probability $p_n$. \footnote{We employ the BERT\textsubscript{base}'s last layer's hidden-state of the first token of the sequence (i.e., pooled CLS embeddings) as $\mathbf{\tilde{x}_n}$, which is then $L_2$ normalized.} A contrastive loss function is then applied in a supervised fashion to maximize the encoder's capability of building downstream optimized sequence representations (see Section \ref{sec:contrastive_learning}). After this contrastive training, the encoder parameters are frozen and a linear classification layer is then trained with cross-entropy. In the remainder of this section, we review the self-supervised contrastive function \citep{Gao2021} and its extended \textit{supervised} counterpart inspired by \citet{khosla2020}.

\subsection{Contrastive Learning}
\label{sec:contrastive_learning}
Let $i \in I \equiv \{1\cdots MN\}$ be the index of all the encoded sequence embeddings $\mathbf{\tilde{X} \equiv \{\tilde{x}_1 \cdots \tilde{x}_{MN}\}}$ in an input batch. Each sample $i$ is forward propagated $N$ time using distinct drop-out masks, generating altered views denoted as $\mathbf{\tilde{x}_{j(i)}} = Enc(\mathbf{x_i}, p_n)$, where $j(i)$ refers to the indices of the altered \textit{view(s)} for the sample $i$ (also called \textit{positive pairs}). In self-supervised contrastive learning~\cite{Gao2021,khosla2020}, the cost function is formulated as:

\begin{multline}\label{Eq:1}
    \mathcal{L}_{i}^{\mbox{self-sup}} \\= - \sum_{i \in I} \mbox{log} \frac{e^{sim\big(\mathbf{\tilde{x}_i}, \mathbf{\tilde{x}_{j(i)}}\big)/\tau}}
    {\sum_{b \in B(i)} e^{sim\big(\mathbf{\tilde{x}_i}, \mathbf{\tilde{x}_{b}}\big)/\tau}},
\end{multline}

Where, $B(i) \equiv I \backslash \{i\} $ is the set of so called \textit{negative} pairs for the anchor $\mathbf{\tilde{x}_i}$. $\tau$ is a temperature scaling parameter. $sim(.)$ stands for any similarity function such as cosine similarity or the inner product.

The main shortcoming of self-supervised contrastive learning is that since the class \textit{labels} of the inputs are ignored, the samples from the same class might end up being employed as the negative pairs (e.g. $B(I)$) and therefore hurt the training performance. For instance, in CoLA~\citep{warstadt2018neural} the aim is to determine whether the input is grammatical or ungrammatical. An unsupervised contrastive learning in this case might employ a grammatically correct sentence as the negative pair for the input anchor (see Figure~\ref{fig:SupCSE}). 

In order to avoid this limitation and make the system able to learn in a \textit{supervised} fashion, \citet{khosla2020} extended Equation~\ref{Eq:1} to account for input labels. Given $M$ annotated samples $\{\tilde{x}_i, \tilde{y}_i\}_{i=1...M}$ passed $N$ times through the \textit{dropout} masks, the \textit{supervised} contrastive learning loss is defined as:

\begin{multline}\label{Eq:2}
    \mathcal{L}_{i}^{\mbox{sup}} = \sum_{i \in I} \frac{-1}{|P(i)|}\\ 
    \sum_{p \in P(i)} \mbox{log} \frac{e^{sim\big(\mathbf{\tilde{x}_i}, \mathbf{\tilde{x}_{p}}\big)/\tau}}
    {\sum_{b \in B(i)} e^{sim\big(\mathbf{\tilde{x}_i}, \mathbf{\tilde{x}_{b}}\big)/\tau}},
\end{multline}

Where $P(i) \equiv \{p \in B(i): \tilde{y}_p = \tilde{y}_i \}$ is the positive pair set distinct from sample $i$ and $|.|$ stands for cardinality (for details on derivation of Equation~\ref{Eq:2} see~\citet{khosla2020}). \textit{SupCL-Seq} employs $\mathcal{L}_{i}^{\mbox{sup}}$ as contrastive loss function.

\begin{table}[t]
\centering
\resizebox{0.8\columnwidth}{!}{
\begin{tabular}{llll}
\textbf{Task}                  & \textbf{Drop-out}         & \textbf{Batch size} & \textbf{Score} \\\hline
\multirow{5}{*}{\textbf{CoLA}} & [0.0,0.1,0.2,0.3,0.4] & 800                 & \textbf{61.2}  \\
                               & [0.0,0.1,0.2,0.3]     & 640                 & 57.9           \\
                               & [0.0,0.1,0.2]         & 480                 & 58.9           \\
                               & [0.0,0.1]             & 320                 & 57.9           \\
                               & [0.1,0.1]                       & 256                 & 60.7           \\
\hline\hline                               
\multirow{5}{*}{\textbf{RTE}}  & {[}0.0,0.1,0.2,0.3,0.4{]} & 800                 &      63.5          \\
                               & {[}0.0,0.1,0.2,0.3{]}     & 640                 & 62.4           \\
                               & {[}0.0,0.1,0.2{]}         & 480                 & \textbf{69.3}  \\
                               & {[}0.0,0.1{]}             & 320                 & 63.8           \\
                               & [0.1,0.1] & 256 & 65.3\\  
\hline
\end{tabular}
}
\caption{Effects of different dropout masks and number of views on CoLA and RTE tasks. Score denotes Matthews Correlation Coefficient.}
\label{tab:dropout}
\end{table}

\section{Experiments}

We performed a set of experiments to i) evaluate the effect of number and level of dropout passes on two challenging datasets (see \ref{sec:dropout}); ii) compare the performance of a standard BERT\textsubscript{base} \cite{BERT2018} architecture with a \textit{SupCL-Seq}-empowered BERT\textsubscript{base} model on several benchmarks in GLUE \citep{wang-etal-2018-glue} (see \ref{sec:glue}); iii) compare the performance of \textit{SupCL-Seq} with the \textit{self-supervised} contrastive approach introduced by \citet{Gao2021} in a subset of tasks (see \ref{sec:supervised}); and, finally, iv) assess whether the improvements achieved with our approach are solely due to augmentation (i.e., \textit{dropout} masks) and to which extend contrastive loss helped (see \ref{sec:augmented}).

\begin{table*}[!ht]
\centering
\resizebox{\textwidth}{!}{
\begin{tabular}{ l c c c c c c c c c}
\textbf{System}  & \textbf{QQP} & \textbf{CoLA} & \textbf{MRPC} & \textbf{RTE} & \textbf{STS-B} & \textbf{SST-2} & \textbf{QNLI} & \textbf{WNLI} & \textbf{MNLI}\\ 
\textbf{Nr. Training Samples}  & 363k & 8.5k & 3.5k & 2.5k & 5.7k & 67k & 108k & 635k & 392k\\ 
\hline\hline
BERT\textsubscript{base} - \textit{Standard} &71.2 &55.2 & 86.6/80.3 & 64.6 & 88.0/87.7 & 92.6 & 90.5 &56.6& 84.1 \\

BERT\textsubscript{base} - \textit{SupCL-Seq}  &\textbf{85.9} &\textbf{61.2} & \textbf{89.7/85.7} & \textbf{69.3} & \textbf{89.3/89.1} & 93.2 & \textbf{91.0} & 56.6 & \textbf{84.5}\\ \hline

BERT\textsubscript{base} - \textit{Dropout Augmented}  & - &60.9& 88.9/84.5 & 63.1 & 80.4/81.6 & \textbf{93.4} & - & - & -\\

\hline
\end{tabular}%
}
\caption{ GLUE Test results. BERT\textsubscript{base} - \textit{Standard} is our implementation using the reported hyper-parameters in~\citet{BERT2018} for each task. BERT\textsubscript{base} - \textit{Dropout Augmented} is the standard version trained also on augmented samples. Matthews Correlation Coefficient is reported for CoLA, Pearson/Spearman correlations for STS-B, F1/Accuracy for MRPC , F1 score for QQP, and
accuracy scores are reported for the other tasks.
\label{table:comparison}}
\end{table*}

\subsection{Dropout Levels}
\label{sec:dropout}
In order to study the effect of the number and the level of dropout passes, we assessed the performance of several configurations of BERT\textsubscript{base} on CoLA \citep{warstadt2018neural} and RTE \citep{10.1007/11736790_9} datasets. \citet{Gao2021} empirically showed that using two distinct dropout masks with the same probability of $p=0.1$ lead to the highest performance in their settings. In our supervised experiments, instead, we can generate \textit{views} with different levels of noise, as the system can always rely on their labels. Therefore we choose different parameters, using intervals of $0.1$ for the dropout probabilities. Table~\ref{tab:dropout} reports the results for both datasets. While clear improvements are visible on \textbf{CoLA} when more masks are applied, experiments on \textbf{RTE} show that this is not always the case. In the latter dataset, in fact, performance fluctuates largely across the settings, achieving the highest score when three passes are used. This suggests that the number and level of dropout passes is a task-dependent hyper-parameter. 

\subsection{GLUE Tasks}
\label{sec:glue}
In order to assess the benefit of \textit{SupCL-Seq}, we compared the performance of a standard BERT\textsubscript{base} architecture with the one of a \textit{SupCL-Seq}-empowered BERT\textsubscript{base} model on numerous tasks from the GLUE benchmark~(for a detailed description of each task see \citet{wang-etal-2018-glue}). GLUE also includes a regression task (i.e., \textbf{STS-B}), which requires no architecture modifications~\footnote{To employ \textit{SupCL-Seq}, we rounded to first decimal digit and grouped by similar labels, employing the Mean Squared Error (MSE) Loss for the baselines.}. For the classification experiments, we deploy the hyper-parameters reported in~\citet{BERT2018}. Appendix~\ref{sec:appendixA} details our grid-search-based training details for \textit{SupCL-Seq}. Results are described in Table~\ref{table:comparison}, rows one and two. As it can be noticed, in all cases the \textit{SupCL-Seq}-empowered BERT\textsubscript{base} model obtains equal or higher performance compared to the standard implementation.

\subsubsection{Is it the \textit{dropout augmentation} or the \textit{loss-function}?}
\label{sec:augmented}
In order to study whether the performance gain observed in the previous experiments is solely due to the \textit{dropout} augmentation, we ran a new set of experiments on the smaller datasets (i.e., MRPC, RTE, STS-B and SST-2) in which the standard BERT\textsubscript{base} is trained also on the augmented samples (for the training parameters, see~Appendix~\ref{sec:appendixA}). Table~\ref{table:comparison}, third row, shows the results. While we notice performance gains compared to the BERT\textsubscript{base} - \textit{Standard} in a few tasks, augmentation does not always help. For example, the score for the augmented row is lower in the \textbf{RTE} dataset. Interestingly, dropout augmentation significantly hurts the performance ($\approx 8$ points) in \textbf{STS-B} dataset, where \textit{MSE} loss is employed. We also observe that for all CoLA, MRPC and STS-B, \textit{SupCL-Seq} outperforms the augmented variant, suggesting that its gains are due to the combination of augmentation and contrastive learning, rather than from only the former.

\begin{table}[!ht]
\centering
\resizebox{\columnwidth}{!}{
\begin{tabular}{l c c c}
\textbf{System} & \textbf{RTE} & \textbf{CoLA} & \textbf{MRPC}\\
\hline
BERT\textsubscript{base} - \textit{Self-supervised-CL}  & 55.6 & 35 & 79/68.3\\
BERT\textsubscript{base} - \textit{SupCL-Seq}  & \textbf{69.3} & \textbf{61.2} & \textbf{89.7/85.7}\\
\hline
\end{tabular}
}
\caption{Comparison of unsupervised and supervised contrastive loss. }
\end{table}

\subsection{\textit{Supvervised} Versus \textit{Unsupervised} contrastive Learning}
\label{sec:supervised}
Since, to the best of our knowledge, the only previous attempt of using contrastive learning for improving sequence representation in NLP was performed by \citet{Gao2021} -- they used a \textit{self-supervised} approach to improve the semantic representation, adopting a loss similar to Equation~\ref{Eq:1} --, in Table \ref{sec:supervised} we compare the performance of a linear layer trained on top of their representations with the one of a linear layer trained on top of our representations, which are instead optimized in a \textit{supervised} fashion while the parameters of the base model are kept frozen. \textit{SupCL-Seq} significantly outperforms the re-implementation of \citet{Gao2021}, with larger gains in non-semantic tasks (e.g. CoLA), suggesting that our representations are optimized for the given downstream tasks.

\section{Discussion and Conclusion}
In this paper, we introduced \textit{SupCL-Seq} a \textit{supervised} contrastive learning framework for optimizing sequence representations for downstream tasks. In a series of experiments, we showed that \textit{SupCL-Seq} leads to large performance gains in almost all GLUE tasks when compared to both a standard BERT\textsubscript{base} architecture and an augmented BERT\textsubscript{base} (i.e., improvements are not only due to augmentation). We also investigated the effect of number and level of \textit{dropout} passes, finding that this has to be treated as a task-dependent hyper-parameter, to be fine tuned in a validation set. Finally, we compared our \textit{supervised} approach to the \textit{self-supervised} method by \citet{Gao2021}, showing consistent performance improvements, especially in non-semantic tasks, where the \textit{self-supervised} approach is weaker. These encouraging results open the door to multi-task learning applications of \textit{SupCL-Seq}, where the optimization needs to be constrained towards multiple objectives.



\section*{Acknowledgments}
We would like to thank the reviewers and the chairs for their insightful reviews and suggestions.

\bibliography{anthology,custom}
\bibliographystyle{acl_natbib}

\newpage

\appendix
\section{Training Details}
\label{sec:appendixA}
\begin{table}[!ht]
\centering
\resizebox{\columnwidth}{!}{
\begin{tabular}{l c c c}
\textbf{Task} & \textbf{Learning Rate} & \textbf{Batch Size} & \textbf{dropout} \\
\hline 
\textbf{CoLA} & $5e-05$ & $128$ & $[0.0, 0.1, 0.2]$ \\ 
\textbf{MRPC} & $1e-4$ & $128$ & $[0.0, 0.05, 0.1,0.2]$\\ 
\textbf{RTE} & $1e-4$ & $48$ & $[0.0, 0.1, 0.2]$ \\ 
\textbf{STS-B} & $1e-4$ & $64$ & $[0.0, 0.05, 0.1, 0.2]$ \\ 
\textbf{SST-2} & $5e-05$ & $320$ & $[0.0, 0.1, 0.2]$ \\ 
\textbf{WNLI} & $1e-04$ & $320$ & $[0.0, 0.1, 0.2]$ \\
\textbf{QNLI} & $5e-05$ & $48$ & $[0.0,0.2]$ \\ 
\textbf{QQP} & $5e-05$ & $16$ & $[0.0,0.2,0.3,0.4,0.5]$ \\ 
\textbf{MNLI} & $5e-05$ & $8$ & $[0.1,0.1]$ \\ 
\end{tabular}
}
\caption{Contrastive learning training details per GLUE task. All of the tasks were trained for $5$ epochs (except \textit{QNLI}, \textit{QQP} and \textit{MNLI} that were trained for $2$, $1$ and $3$ epochs respectively) and $\tau = 0.05$.}
\label{tab:traindetails}
\end{table}

\textit{SupCL-Seq} is implemented on top of the Huggingface's trainer python package~\cite{DBLP:journals/corr/abs-1910-03771}\footnote{\href{https://github.com/huggingface/transformers}{https://github.com/huggingface/transformers}}. For the $sim(.)$ (similarity) function, we employed \textit{inner dot product}. For supervised contrastive learning, we employed the hyperparameters detailed in Table~\ref{tab:traindetails}. We used a grid search strategy for our hyperparameter optimization, where the number of dropouts and their corresponding probability were set to two (i.e. $[0.1,0.1]$) and five respectively ($[0.0,0.1,0.2,0.3,0.4]$). For the learning rate we employed a range of $[5e-05, 1e-4]$.

\end{document}